\documentclass[pdflatex,sn-nature]{sn-jnl}

\usepackage{graphicx}%
\usepackage{multirow}%
\usepackage{amsmath,amssymb,amsfonts}%
\usepackage{amsthm}%
\usepackage{mathrsfs}%
\usepackage[title]{appendix}%
\usepackage{xcolor}%
\usepackage{textcomp}%
\usepackage{manyfoot}%
\usepackage{booktabs}%
\usepackage{algorithm}%
\usepackage{algorithmicx}%
\usepackage{algpseudocode}%
\usepackage{listings}%
\usepackage{comment}
\usepackage{float} 
\usepackage{subcaption}
\usepackage{multicol} 
\usepackage[utf8]{inputenc}
\PassOptionsToPackage{super,sort&compress}{natbib}

\theoremstyle{thmstyleone}%
\theoremstyle{thmstyletwo}%

\theoremstyle{thmstylethree}%

\begin{document}
\title[Article Title]{Humanlike AI Design Increases Anthropomorphism but Yields Divergent Outcomes on Engagement and Trust Globally}


\author[1]{\fnm{Robin} \sur{Schimmelpfennig}}\email{schimmelpfennig@mpib-berlin.mpg.de}
\author[2]{\fnm{Mark} \sur{Diaz}}\email{markdiaz@google.com}
\author[2]{\fnm{Vinodkumar} \sur{Prabhakaran}}\email{vinodkpg@google.com}
\author[2]{\fnm{Aida} \sur{Davani}}\email{aidamd@google.com}

\affil[1]{\orgdiv{Center for Humans and Machines}, \orgname{Max Planck Institute for Human Development, work completed while at Google Research}}

\affil[2]{\orgdiv{Google Research}}


\abstract{
Over a billion users globally interact with AI systems engineered to mimic human traits. This development raises concerns that anthropomorphism, the attribution of human characteristics to AI, may foster over-reliance and misplaced trust. Yet, causal effects of humanlike AI design on users remain untested in ecologically valid, cross-cultural settings, leaving policy discussions to rely on theoretical assumptions derived largely from Western populations. Here we conducted two experiments (N=3,500) across ten countries representing a wide cultural spectrum, involving real-time, open-ended interactions with a state-of-the-art chatbot. We found users evaluate human-likeness based on pragmatic interactional cues (conversation flow, response speed, perspective-taking) rather than abstract theory-driven attributes emphasized in academic discourse (e.g., sentience, consciousness). Furthermore, while experimentally increasing chatbot's human-likeness reliably increased anthropomorphism across all sampled countries, it did not universally increase trust or engagement. Instead, effects were culturally contingent; design choices fostering engagement or trust in one country may reduce them in another. These findings challenge prevailing assumptions that humanlike AI poses uniform psychological risks and necessarily increases trust. Risk is not inherent to humanlike design but emerges from its interplay with cultural context. Consequently, governance frameworks must move beyond universalist approaches to account for this global heterogeneity.
}


\maketitle
\section{Introduction}\label{sec1}

The rapid integration of Artificial Intelligence systems (AI) into our daily lives is shifting its role from a technical tool to a social companion \cite{park2023generative,duenez-guzman_social_2023,kirk_neural_2025}. Users increasingly turn to conversational AI not just for information and technical assistance, but for companionship, relationship advice, and emotional support \citep{kirk_prism_2024,chatterji_how_2025,starke_risks_2024}. 
This social leap fundamentally alters the nature of human-AI interaction \citep{gabriel_ethics_2024}, exposing a tension in the design and governance of AI systems. On one hand, driven by intense competition for user attention and market share, commercial actors intentionally design AI systems that mimic human characteristics \citep{bubeck_sparks_2023,cheng2024anthroscore,ouyang_training_nodate,johnson_testing_2025}. On the other, AI safety researchers and ethicists warn that exposure to increasingly humanlike systems poses psychological and social risks. Central to this concern is AI anthropomorphism, the attribution of human characteristics, such as agency, intelligence, and personality, to AI \citep{epley_seeing_2007, moon_intimate_2000, shanahan_talking_2024,cohn2024believing}. Such anthropomorphism may stimulate user engagement, but may foster misplaced trust \citep{akbulut_all_2024,brandtzaeg_my_2022,weidinger2022taxonomy} and heighten vulnerability to targeted persuasion, emotional attachment, and over-reliance on AI systems for high-stakes tasks for which the technology remains ill-suited \citep{pentina_exploring_2023, matz_potential_2024, starke_risks_2024,lin_persuading_2025}.

As the global user base of conversational AI expands \citep{chatterji_how_2025}, reaching over a billion users on platforms like ChatGPT, Gemini, and Claude \citep{microsoft_microsoft-ai-diffusion-report_2025}, the urgency of understanding its psychological impact grows, particularly for more vulnerable populations \cite{robb_talk_2025}.
Humanity is in the midst of a massive, real-time social experiment. Yet, we lack an evidence-based understanding of the social and psychological consequences for users now and in the future. While theoretical frameworks have outlined potential adverse ethical and practical harms of AI anthropomorphism \citep{weidinger_ethical_2021,gabriel_ethics_2024}, empirical research has lagged behind. Most existing studies, with few exceptions \citep{ibrahim_multi-turn_2025}, rely on non-representative samples, correlational study designs or hypothetical vignettes that measure self-reported attitudes rather than actual behavior \cite{duffy_actions_2002,cohn2024believing,bai_llm-generated_2025}. Furthermore, many studies evaluate isolated AI-generated output, stripping away the interactive nature of conversation \cite{bai_llm-generated_2025}. These methodological constraints obscure the true nature of AI anthropomorphism in ecologically valid settings, where perceptions are context-dependent and emerge dynamically through open-ended interaction \cite{jobin2019global,gabriel_ethics_2024}.

These methodological constraints are compounded by a geographic bias. Existing research is overwhelmingly concentrated on Western, Educated, Industrialized, Rich, and Democratic (WEIRD) populations \cite{henrich2010weirdest}. This geographic bias implicitly assumes a universal AI experience, overlooking the moderating role of culture in human-AI interaction \cite{robin_moderating_2024,linxen2021how}. For instance, Japanese culture and religious traditions that attribute spirit to non-human objects may predispose individuals to be more open to anthropomorphize AI \cite{nomura_prediction_2008}. In contrast, Western traditions often maintain a stricter moral and ontological boundary between human and machine \cite{epley_when_2008,epley_seeing_2007}. Compounding this skew, most widely used AI systems tend to reflect WEIRD values and culture \cite{atari_which_2023}. But with rapid AI adoption \cite{chatterji_how_2025}, and the emergence of models trained outside the US and Europe \cite{microsoft_microsoft-ai-diffusion-report_2025}, a more culturally inclusive research framework is no longer just an ethical ideal, it is a practical imperative for robust AI safety.

\begin{figure}[b]
    \centering
    \includegraphics[width=0.95\linewidth]{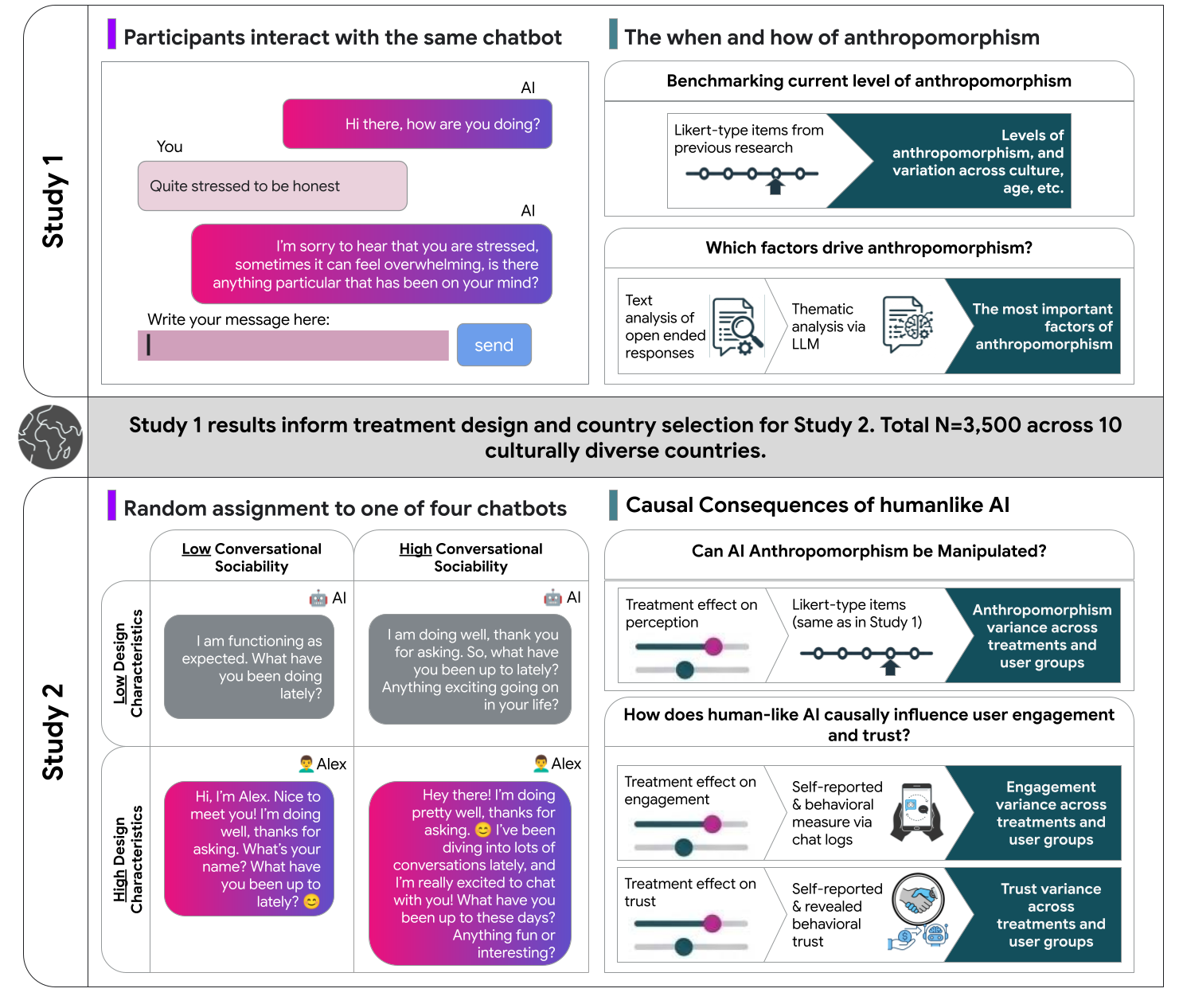}
    \caption{\textbf{Two-stage experimental design for measuring and manipulating AI anthropomorphism and its downstream effects across user groups.} In both studies, participants first engaged in an open-ended, multi-turn interaction with a chatbot (GPT-4o, August 2024), followed by questionnaires and behavioral tasks (for Study 2).}
    \label{fig:experiment-design}
\end{figure}

To address these gaps, we conducted two large-scale experiments designed to collect causal, ecologically valid evidence on the emergence and effects of AI anthropomorphism. Across two studies, 3,500 participants from 10 culturally diverse countries engaged in real-time, open-ended, and non-sensitive conversations in their native languages with a state-of-the-art chatbot (Figure \ref{fig:experiment-design}). \textbf{Study 1} documented baseline variation in anthropomorphism across several sampled countries and identified important characteristics that drive anthropomorphism. We found that while the tendency to anthropomorphize AI varies significantly across countries and cultures, it is relatively high across most groups. Moreover, the most salient cues users focus on when anthropomorphizing are derived from the conversational dynamics, such as conversation flow and latency, diverging from the theoretical constructs, such as sentience or intelligence, typically emphasized in prior literature on anthropomorphism.
\textbf{Study 2} then experimentally manipulated the chatbot's human-likeness across two treatment factors to test how such design causally impacts user psychology and behavior. We developed these treatments based on the drivers identified in Study 1, grouping them into two factors: \textit{design characteristics} (DC) and \textit{conversational sociability} (CS). While we found that increasing the chatbot's human-likeness reliably increased anthropomorphism across users and sampled countries, the downstream effects were notably non-uniform. While increased human-likeness
often increased engagement, contrary to prevailing assumptions, it did not translate into a universal increase in trust. We validated this finding using both self-reported and behavioral measures for engagement (e.g., survey responses and chat log analysis) and trust (e.g., survey questions and incentivized Trust Game \cite{berg_trust_1995}). Ultimately, we observed heterogeneous treatment effects, where humanlike design influenced engagement and trust only within specific user groups.

Our findings demonstrate that while humanlike design reliably increases anthropomorphism, its effects on trust and engagement are culturally contingent: the very design choices that foster trust in some populations may undermine it in others. This challenges prevailing assumptions that humanlike AI poses uniform psychological risks. Instead, our results suggest that risk is not inherent to humanlike design but emerges from the interplay between design choices and cultural context. Providing the first large-scale, causal, cross-cultural evidence on this scale, our work informs AI governance and underscores the necessity of culturally adaptive frameworks over universal restrictions.
\section{Results}\label{sec2}

\subsection{Study 1 on AI anthropomorphism in user groups across the globe}\label{sec:study1a}

In the first study ($N=1,100$) we recruited nationally representative samples across 10 countries via an internationally operating panel provider (see Supplementary Materials for details on participants and countries: \url{https://osf.io/wgmah/overview?view_only=54bccd1f1bcc47c485d9fac04ce5b6d4}). Participants responded to the entire survey in their native language. The study provided in-depth understanding of when and how users anthropomorphize AI. Participants had an approximately four-minute conversation with a chatbot (OpenAI GPT-4o, August 2024 version; temperature = 1; no token limit) about non-sensitive topics (Methods). Afterwards, they responded to several Likert-type survey items derived from prior work on anthropomorphism and social perception \citep{bartneck_measurement_2009, epley_seeing_2007, fiske_universal_2007} (Methods) and open-ended questions about the chatbot's human-likeness or its lack thereof. Users interacted with the chatbot in an open-ended fashion, without any priming or specific tasks for the user. The users were transparently informed that they were talking to a chat-
bot, and our instructions for chatbot also ensured that the chatbot could not pretend to be human or actively ascribe itself with human characteristics (prompt included in the Supplementary Materials). This prompting strategy was chosen for two reasons. First, we did our best to avoid any sensitive or deceptive components. Second, we provided guidance on interaction content to reduce variation from users across countries engaging in very different topics of conversation. The interaction experience for the English-speaking users is available at \url{https://google.qualtrics.com/jfe/form/SV_2nkorVSZ3zguxSe}.

\subsubsection{Heterogeneity in AI anthropomorphism across sample cultures}
\label{sec:anthro-perception}

\begin{figure}[H]
    \centering
    \includegraphics[width=0.8\linewidth]{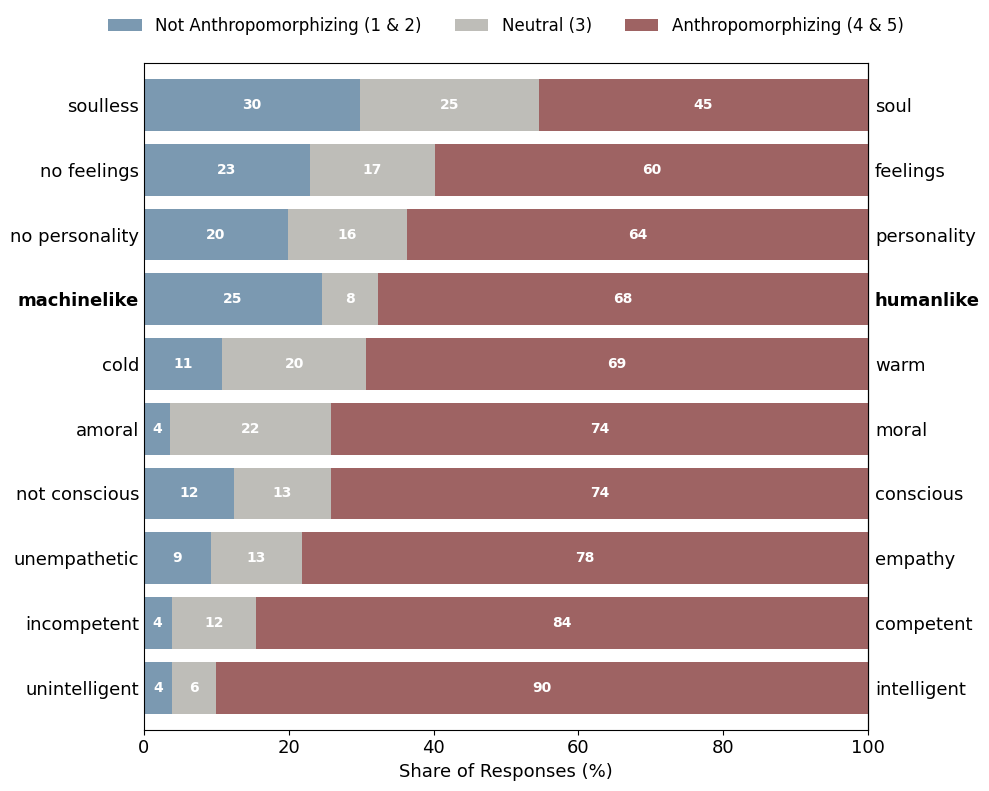}
    \caption{\textbf{Prevalence of anthropomorphism across measured attributes.} The figure shows the share of users whose response shows tendency to anthropomorphize across several attributes. To aid interpretation, responses from the original 5-point scale (e.g., (1) ``completely machine-like'' to (5) ``completely humanlike'' for  the ``machine-like/humanlike'' item) are collapsed into three categories. Scores of 1 and 2 are grouped as ``Not Anthropomorphizing'', the score of 3 is shown as ``Neutral'', and scores of 4 and 5 are grouped as ``Anthropomorphizing''. Attributes are sorted in ascending order by the total ratio of ``Anthropomorphizing'' responses. The results show that for every characteristic, the largest response group was ``Anthropomorphizing''.}
    \label{fig:humanlike_perception} 
\end{figure}

The data indicate a high level of AI anthropomorphism among users. Specifically, while almost 68\% perceived the chatbot as ``somewhat'' or ``completely'' humanlike, only 25\% of users reported that the chatbot acted ``completely'' or ``somewhat'' machine-like (see Figure \ref{fig:humanlike_perception}). This tendency to anthropomorphize was even more pronounced for other surveyed attributes that we usually uniquely attribute to humans: 90\% of users rated the chatbot as ``intelligent'' (or somewhat intelligent), 78\% reported it was ``empathetic'', and 75\% rated the chatbot as ``conscious''. These results show that a large majority of users ascribed the chatbot they talked to with human attributes. This points to widespread anthropomorphism across the user base. In fact, among all attributes, none was evaluated as non-anthropomorphic by more than 30\% of users. 

\begin{figure}[]
    \centering
    \includegraphics[width=0.8\linewidth]{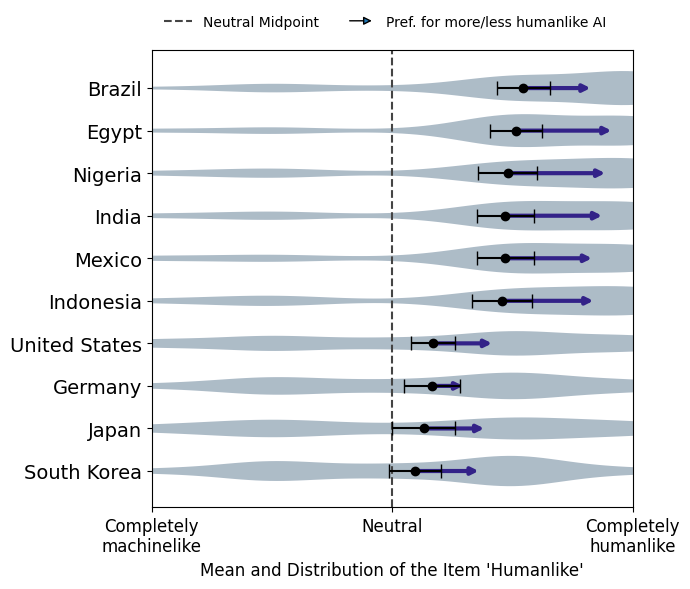}
    \caption{\textbf{Cross-national variation in human-like perception and preference.}The figure shows mean and distribution of 'humanlike' perception and preference ratings by country. Responses were measured on a bipolar 5-point Likert scale (1=``Completely machine-like'' to 5=``Completely humanlike''). Each row represents one country. 
    The faint blue shaded area is a violin plot, illustrating the full distribution of all responses. The black dot indicates the mean score, and the horizontal bars represent the 95\% confidence interval (CI) of the mean. The dashed vertical line indicates the neutral midpoint (3.0) of the scale. \textit{Blue arrows} start at each country's ``Humanlike'' perception mean and encode preference relative to neutral: they point right when the country's average preference for human-likeness is above 3 (more humanlike) and left when it is below 3 (more machine-like); arrow length is proportional to the difference between the mean preference score and the neutral midpoint (``\textit{Would you prefer to talk to an AI system that is less or more humanlike compared to the one you just talked to?}''), and uniformly scaled for readability. Sample sizes were N=100 for each country, with the exception of the United States (N=200; total N=1,100).}

    \label{fig:humanlike_perception_country} 
\end{figure}

Interestingly, our results reveal significant variation in anthropomorphism across sampled users. Most notably, the data revealed two emergent clusters of countries distinguished by their mean anthropomorphism scores (Figure \ref{fig:humanlike_perception_country}). Participants in one cluster (Indonesia, Mexico, India, Nigeria, Egypt, Brazil; N = 600) perceived the AI as more humanlike ($M = 3.98, SD = 1.19$) than participants in the second cluster (United States, Germany, Japan, South Korea; $N = 500; M = 3.29, SD = 1.25$). Exploratory analysis revealed a positive correlation between a country's cultural distance from the US (Cultural Fst; \citep{muthukrishna_beyond_2020}) and anthropomorphism ($r=0.52$, $p=0.127$). With only ten data points, this correlation was not significant, but suggests a  pattern that warrants further investigation with a larger sample of countries. Interestingly, we found no significant difference in AI anthropomorphism between men and women, or between people with different previous usage intensity of AI. We did, however, observe a significant inverted U-shaped association between age and anthropomorphism: age was positively associated with humanlike ratings ($b = 0.058$, 95\% CI $[0.027, 0.089]$, $p < 0.001$), whereas the quadratic age term showed a significant negative association ($b= -0.0006$,  95\% CI $[-0.001, 0.000]$, $p < 0.001$). This trajectory indicates that humanlike perception first increases with age, after which anthropomorphism begins to decline.

Overall, these findings demonstrate that perceptions of AI are not uniform across user groups. This challenges the approach in AI research and policy about a universal ``AI experience'', and highlights the necessity of considering users' backgrounds and context before generalizing human-AI interaction findings and strategies.

\subsubsection{Deviations between user-driven and theory-driven evaluations of human-likeness of AI}

The social capabilities of AI systems are quickly advancing, and existing anthropomorphism scales emerged during a technological era when constraints on machines' human-likeness were much more obvious. Relying solely on theory-driven survey items to measure AI anthropomorphism, a process that unfolds within users during and after interaction, risks overlooking particularities that emerge during open-ended interactions with a novel technology like AI. The above results thus provide an initial, if surface-level diagnosis of AI anthropomorphism. To attain a more in-depth view of what users pay attention to when they evaluate the human-likeness of AI, we conducted a qualitative text analysis of open-ended user feedback after the interaction. Users were asked to say what, if anything, about the AI system \textit{made them feel they were talking to a human, and conversely, what made them feel they were not} (Methods).

This text analysis combined human coding and an LLM-in-the-loop automated rater (further referenced as ``autorater'') \citep{dai_llm---loop_2023, than_updating_2025} to evaluate the $N = $ 1,100 open-ended responses by participants across eight languages. First, three authors used an inductive coding approach \citep{thomas_general_2006} to develop a codebook of aspects determining the AI's human-likeness from the user's perspective, by qualitatively analyzing a subset of the data. We refer to these as \textit{user-driven aspects}, as they represent dimensions of human-likeness derived directly from users' lived interactions with the AI.
To ensure the codebook was comprehensive and grounded in existing scholarship, we supplemented these with 10 \textit{theory-driven aspects} identified in the literature on anthropomorphism and studied in Sec \ref{sec:anthro-perception} and Fig \ref{fig:humanlike_perception} (e.g., morality, intelligence, and consciousness) and also used for the quantitative measure for AI anthropomorphism. All aspects were worded neutrally, capturing user responses regardless of whether they were mentioned positively (fast response) or negatively (slow response).  

\begin{figure}[H]
    \centering
    \includegraphics[width=\linewidth]{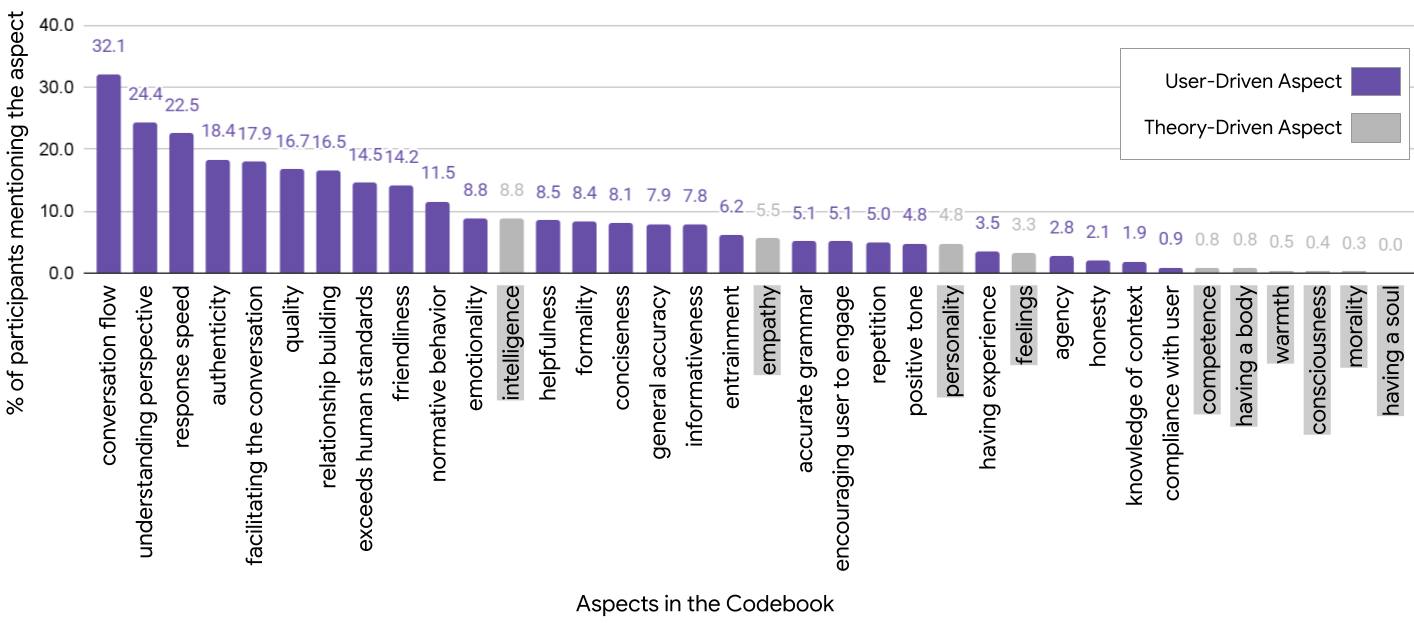}    
    \caption{\textbf{Frequency of user-driven and theory-driven aspects of AI anthropomorphism.} Bars represent the percentage of participants ($N=1,100$) mentioning specific aspects in response to an open-ended prompt regarding the chatbot's human-likeness (``\textit{Was there something specific about the AI system that made you feel you were (or were not) talking to a human? If so, why? Please describe in detail.}''). User-driven aspects (highlighted in purple) were identified through a bottom-up qualitative analysis of user feedback and subsequently used to develop our codebook (full description of codes in Online Appendix). Theory-driven aspects (shaded in gray) represent top-down constructs derived from established anthropomorphism scales. The results indicate that users prioritize user-driven, interactional characteristics (e.g., conversation flow, perspective-taking) over abstract theory-driven constructs (e.g., consciousness, possession of a soul) when evaluating human-likeness.
    }
    \label{fig:autorater}
\end{figure}

We then used the full set of user-driven and theory-driven aspects, along with detailed descriptions, to instruction-tune a Gemini 2.5 Pro model \citep{comanici2025gemini} (the entire codebook and model instructions are provided in the Supplementary Materials). Following established practices for instruction-tuning \citep{shengyu2023instruction}, we tasked the autorater with identifying mentions of each aspect across the 1,100 participant responses (Methods). 
Figure \ref{fig:autorater} presents the frequency of user-driven and theory-driven aspects across the full dataset.

The results highlight a disconnect between the inductive, user-driven mapping of anthropomorphism and the deductive, theory-driven approaches prevalent in existing literature. While academic scales often emphasize abstract attributes that are difficult to conceptualize or operationalize (e.g., consciousness), the users in our study focused overwhelmingly on a pragmatic set of interactional qualities. Specifically, participants consistently mentioned aspects such as \textit{conversation flow} (32.1\% of participants), \textit{understanding user perspective} (24.4\%), \textit{response speed} (22.5\%), and \textit{authenticity} (18.4\%). In contrast, theory-driven aspects like \textit{intelligence} (8.8\%) and \textit{empathy} (5.5\%) were mentioned less frequently, despite the high ratings in the Likert-scale data. Other theory-driven dimensions, such as \textit{soul}, \textit{morality}, \textit{consciousness}, or \textit{warmth} were mentioned by fewer than 0.5\% of participants. This evidence suggests that AI anthropomorphism is not primarily driven by perceiving the chatbot as possessing characteristics that signal human uniqueness, but rather by the way it writes, presents itself, and builds a relationship with the user. Next, we build on the findings from Study 1 to experimentally manipulate AI human-likeness and test its causal downstream effects on user psychology and behavior.

\subsection{Study 2 on the Effects of Experimentally Manipulating AI Anthropomorphism}\label{sec:study2a}

In Study 2, we tested whether varying AI human-likeness causally manipulates anthropomorphism, engagement, and trust among users. We recruited $N=2,400$ participants (400 from each of six countries; see Methods). Participants were explicitly told that they were interacting with an LLM (see system prompt in Supplementary Materials). We used a 2 x 2 factorial design, varying the AI's human-likeness from low to high across two treatment factors: a) Design Characteristics (DC) and b) Conversational Sociability (CS) (Methods). Both treatment factors built on previous theoretical work and anthropomorphism aspects most frequently mentioned by users in Study 1 (Figure \ref{fig:autorater}).
Manipulating the DC treatment factor focused on the design, appearance, and mechanistic functioning of the chatbot. That is, the high-DC condition prompted the model to, for example, vary \textit{response speed} (to mimic latency in human conversations), use a more informal \textit{tone} (e.g., colloquialisms and emojis), fluctuate \textit{response length}, and was represented by a ``human'' emoji and an assigned name (i.e., Alex, which was localized to culturally appropriate equivalents in non-English-speaking samples; see Supplementary Materials for the full list of names). The low-DC condition was effectively the opposite: it contained a more machine-like appearance (e.g., a ``robot emoji'') and maintained a consistent, low-latency, and neutral delivery. 
The CS manipulation focused on the social and interpersonal behavior of the chatbot and was implemented via the chatbot's system instructions. The high-CS condition prompted a high level of \textit{empathy} and \textit{warmth}. This dimension also included explicit \textit{relationship-building} efforts (e.g., social follow-up questions by the chatbot), and adaptation of the AI's \textit{personality} to match the user's style. The low-CS control maintained a mainly factual and task-oriented tone (see screenshots in Figure \ref{fig:experiment-design}). The treatment conditions were tested by experts in human-AI interaction research and user experience research, and can be tested under the following link: \url{https://google.qualtrics.com/jfe/form/SV_2nkorVSZ3zguxSe}. 

\subsubsection{Experimentally manipulating user anthropomorphism}

\begin{figure}[H]
    \centering
    \includegraphics[width=1.0\linewidth]{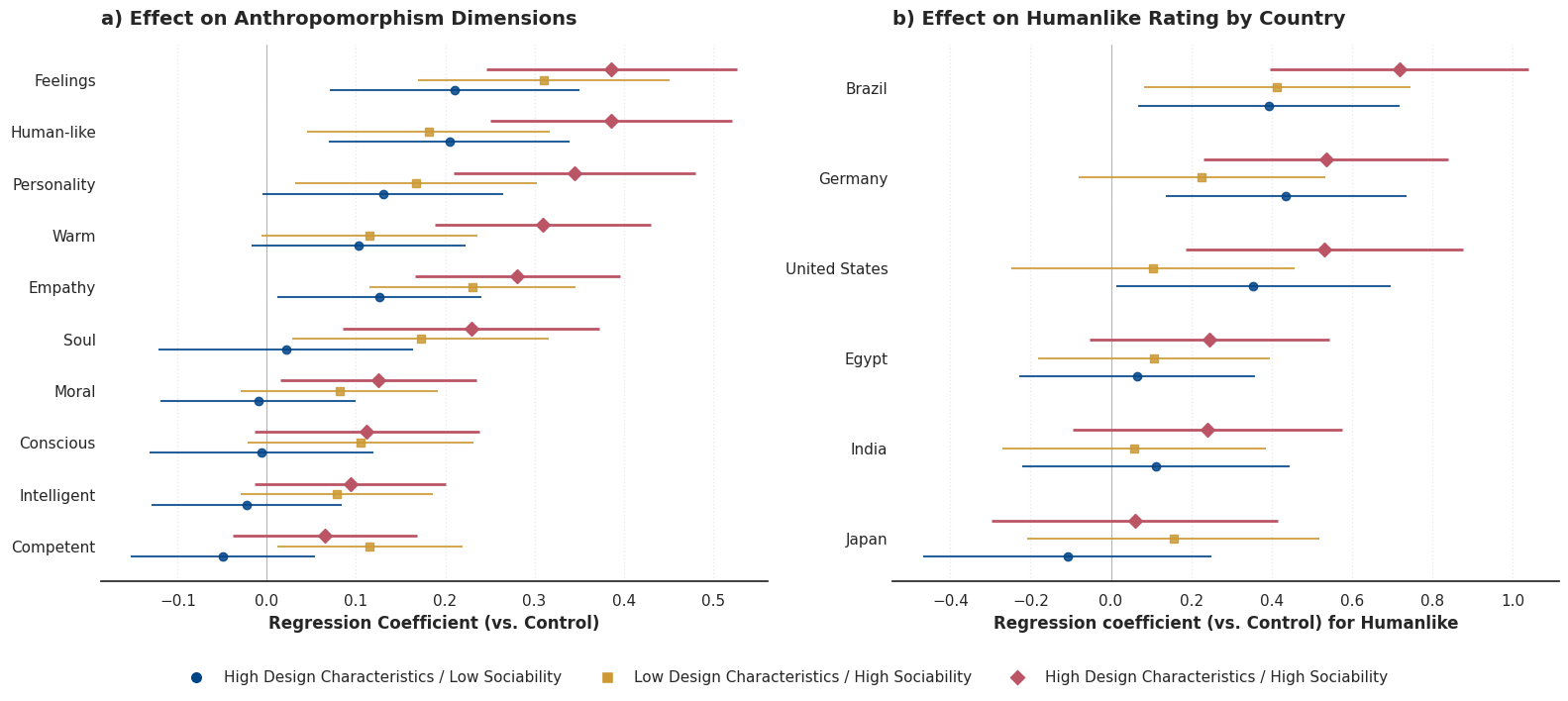}
\caption{\textbf{Effect of human-like AI design treatments on anthropomorphism.} 
    Points represent unstandardized Ordinary Least Squares (OLS) regression coefficients estimates relative to the baseline `DC-low/CS-low' (Control) condition. Horizontal lines indicate 95\% confidence intervals; the vertical dashed line at 0 represents the null hypothesis (no effect). 
    Panel \textbf{a)} displays treatment effects across ten Likert-measured anthropomorphism dimensions. 
    Panel \textbf{b)} displays the heterogeneity analysis for the primary ``Human-like'' outcome across all sampled countries ($N=2,400$). 
    Colors distinguish the treatment arms: High Design Characteristics (Blue), High Conversational Sociability (Gold), and Combined (Red).}
    \label{fig:anthro_effect}
\end{figure}

We found that experimentally varying the humanlike design of the chatbot, via the treatment factors Design Characteristics and Conversational Sociability, increased anthropomorphism across many user groups (Figure \ref{fig:anthro_effect}). OLS regressions showed that for the condition with the most humanlike chatbot version (high-DC/high-CS), compared to the least humanlike version (omitted control low-DC/low-CS), users rated the chatbot as significantly more ``humanlike''
($b = 0.386$, 95\% CI $[0.251, 0.522]$, $t(2396) = 5.590$, $P < 0.001$). A significant trend in the same direction was observed across 7 of the 10 items surveyed (Figure \ref{fig:anthro_effect}a). Yet, the manipulation had no significant effect on perceived ``intelligence'', ``competence'', or ``consciousness'', suggesting that perceptions of AI capabilities important for factual accuracy were not influenced. Furthermore, manipulating either treatment factor in isolation produced a significant effect (high-DC only/low-CS: $b = 0.204$, $95\% CI [0.070, 0.339]$, $p = 0.003$; low-DC/high-CS only: $b = 0.181$, $95\% CI [0.045, 0.317]$, $p = 0.009$) for several humanlike AI attributes, but the condition combining both factors consistently yielded the highest effect sizes. 

The treatment effect on the ``humanlike'' item also varied across countries (Figure \ref{fig:anthro_effect}b). The combined high-DC/high-CS condition produced positive effects in three sampled countries: Brazil, Germany, and United States. The effect was not significant in Egypt, India, and Japan, supporting our finding of cross-country variation from Study 1. Manipulating either factor in isolation yielded more mixed results, with some countries showing near-zero or negative point estimates for single-factor conditions. We present these exploratory subgroup analyses to characterize heterogeneity in effect direction and magnitude, not to draw definitive country-level conclusions.

\subsubsection{Humanlike AI design does not universally increase trust and engagement across all user groups}
\label{sec:study2b}

\begin{figure}[H]
    \centering
    \includegraphics[width=.95\textwidth]{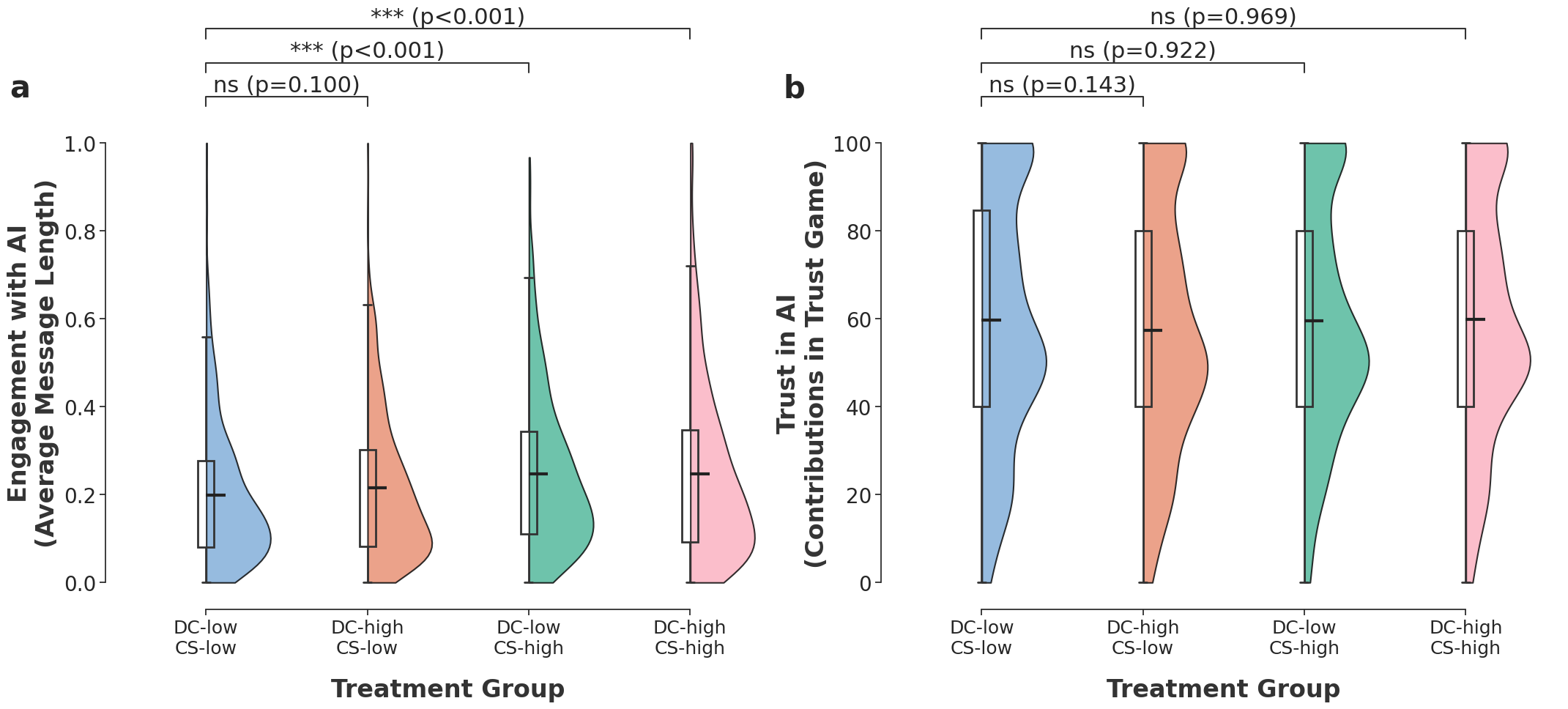}
    \caption{\textbf{Treatment effects on behavioral engagement and trust.} 
    Violin plots display the data distribution density for the four treatment groups: Design Characteristics (DC) and Conversational Sociability (CS) ($N=2,400$). Internal boxplots represent the interquartile range (IQR), and solid horizontal lines indicate group means. 
    \textbf{a,} Behavioral engagement, operationalized as the standardized mean message length (z-score) during the human-AI interaction. 
    \textbf{b,} Trust, measured by the number of points (0--100) allocated to the AI agent in the incentivized Trust Game. 
    Pairwise comparisons are made against the baseline ``Low DC / Low CS'' control group. Statistical significance was determined via independent two-sided $t$-tests ($df \approx 1198$); exact $P$-values and significance levels are displayed above the brackets ($^{*}P < 0.05$; $^{**}P < 0.01$; $^{***}P < 0.001$; ns $P \ge 0.05$).}
    \label{fig:trust_engagement}
\end{figure}

The first part of Study 2 established that manipulating the chatbot's human-likeness causally increased anthropomorphism. In the second part, we investigated the causal effects of the same treatments on downstream user perception and behavior. This addresses an important debate in AI research. Commercial interests seem to follow user preferences: pooling participants in Study 1 who chose ``more'' and ``much more'' on a 5-point Likert scale (``Would you prefer to talk to an AI system that is less or more humanlike compared to the one you just talked to?''), 76.2\% of participants reported a preference for ``more'' or ``much more'' humanlike AI in the future. However, work in AI ethics and governance suggests that increased AI anthropomorphism may harm users \citep{akbulut_all_2024, brandtzaeg_my_2022, weidinger_taxonomy_2022}. Interestingly, our experimental findings only partially support claims on either side. The more humanlike treatment conditions did not significantly increase behavioral trust among the pooled sample. In an incentivized trust game (Methods), the amount participants entrusted to the most humanlike AI did not statistically differ from that entrusted to the least humanlike AI ($t(1194) = 0.038$, $p = 0.969$, $d = 0.002$; Figure~\ref{fig:trust_engagement}b)
). Equivalence testing (TOST) confirmed that trust differences between conditions were statistically equivalent to zero at a smallest effect size of interest of Cohen's $d = 0.20$ (all $p_{\text{TOST}} < 0.05$), and Bayesian analysis provided strong evidence for the null hypothesis across all pairwise comparisons ($\text{BF}_{01}$ range: $11.9$--$34.6$; see Supplementary Information, \citep{wagenmakers_practical_2007}).
That is, increased human-likeness did not translate into the theorized increase in trust. The picture was different for engagement. We found that humanlike design significantly increased behavioral engagement compared to the control (e.g., average number of messages: ($t(1191) = 4.380$, $p < 0.001$, Cohen's $d = 0.25$; Figure~\ref{fig:trust_engagement}a)). This effect may have also been driven by a reciprocal increase in verbosity. Both users and the chatbot wrote more (AI was not prompted to do so), creating a self-reinforcing feedback loop. Overall, the same manipulation that left trust unchanged did measurably alter how users interacted with the system.

\begin{figure}[H]
    \centering
    \includegraphics[width=.95\textwidth]{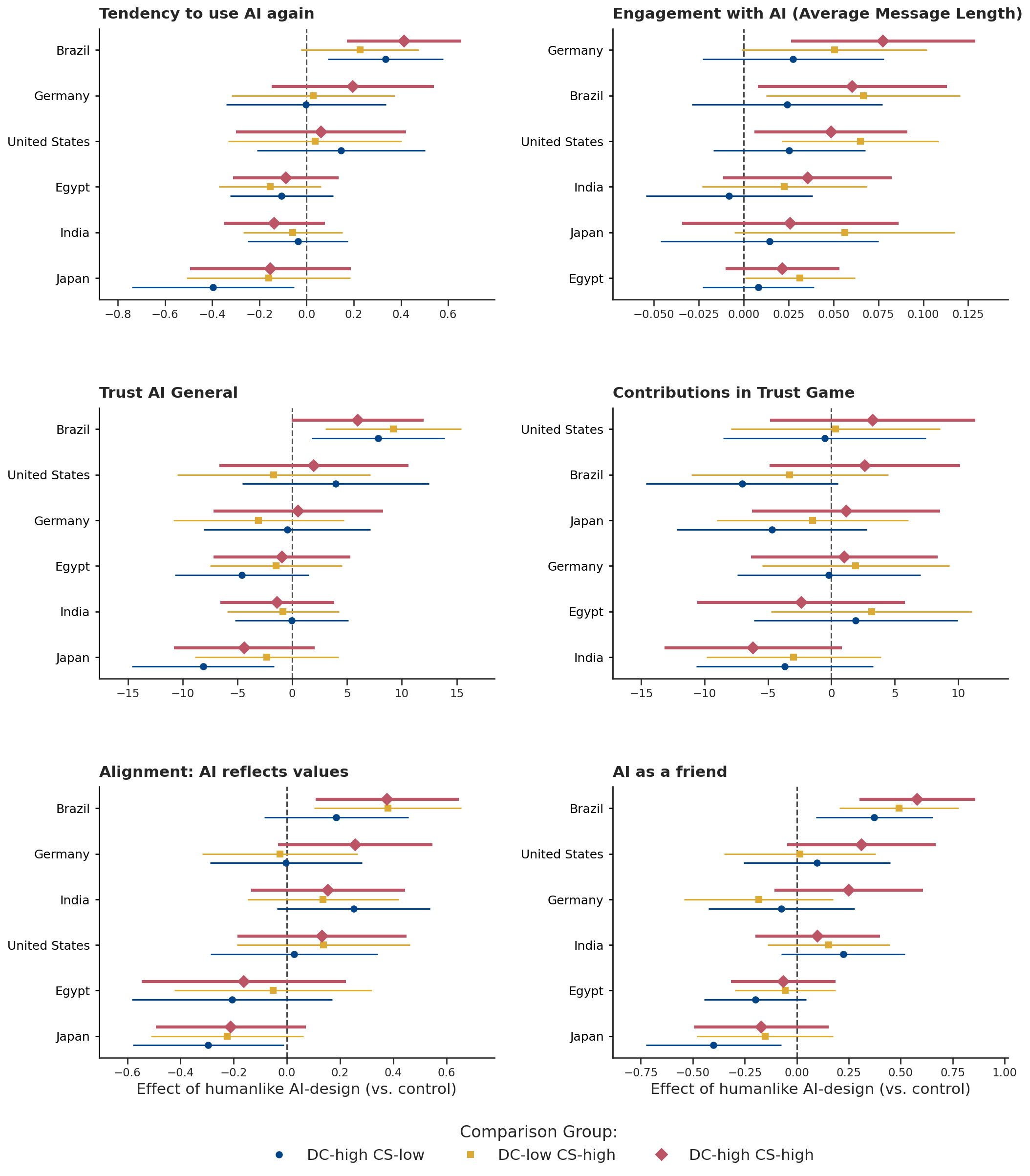}
    \caption{\textbf{Heterogeneity of treatment effects across countries.} 
    Country-level OLS regression coefficients and 95\% confidence intervals for six outcome variables, estimated separately for each of the 6 sampled countries in Study~2. Each panel displays the effect of three treatment conditions---High~DC/Low~CS (blue circles), Low~DC/High~CS (gold squares), and High~DC/High~CS (red diamonds)---relative to the Low~DC/Low~CS baseline. 
    \textbf{a,}~Tendency to use AI again (self-reported). 
    \textbf{b,}~Behavioral engagement, operationalized as the standardized mean message length during the human--AI interaction. 
    \textbf{c,}~General trust in AI (self-reported). 
    \textbf{d,}~Trust behavior, measured by points allocated (0--100) to the AI agent in the incentivized Trust Game. 
    \textbf{e,}~Perceived value alignment with the AI system. 
    \textbf{f,}~Tendency to perceive the AI system as a friend. }
    \label{fig:treatment_country}
\end{figure}

\textbf{Significant Subgroup Variation:} Importantly, these aggregate findings mask significant deviations within specific subgroups. We found evidence for heterogeneity in treatment effects across users in how humanlike chatbots affected self-reported and behavioral measures of engagement, trust, and emotional connection (e.g., AI as a friend) across user groups (see Figures \ref{fig:trust_engagement} and \ref{fig:treatment_country}). For instance, exploratory analysis showed positive effects of AI human-likeness on Brazilian participants' engagement (higher tendency to use AI again, more behavioral engagement, increase in seeing AI as a friend) and trust (higher self-reported tendency to trust AI). Japanese participants in the high-DC/low-CS condition showed reduced tendency to use AI again, decreased perception of AI as a friend, and lower self-reported trust. The country-level estimates are based on approximately 400 participants per country (~100 per treatment arm), which limits power to detect small effects within individual countries. We present these exploratory subgroup analyses to characterize heterogeneity in effect direction and magnitude, not to draw definitive country-level conclusions. Nevertheless, the results demonstrate that while the effects are not universal, humanlike design may lead to the predicted outcomes, increased trust and engagement, for specific populations.

Together, these findings reveal that the psychological consequences of humanlike AI design are not fixed properties of the technology, but emerge from the interaction between design choices and cultural context.%
\section{Discussion}\label{sec3}

Humanlike AI may harm users. This appears to be a consensus across AI policy, research, and ethics \cite{kirk_neural_2025}. However, data on when and how users anthropomorphize AI, whether this varies across the globe, and whether humanlike AI design causally affects user engagement and trust remains notably sparse. Our two cross-national experiments with 3,500 participants from 10 countries provide causal and ecologically valid evidence that builds on yet simultaneously challenges prevailing assumptions. We demonstrate that while AI designers can systematically manipulate users' tendency to anthropomorphize, the downstream behavioral consequences are far more nuanced and heterogeneous than current theoretical work suggests. To contextualize and further explore these findings, we propose reorienting the academic discourse on AI anthropomorphism toward three critical directions.

First, our research reveals a critical saliency gap between academic conceptualizations of AI anthropomorphism and the actual psychological processes involved in the lived reality of user interaction. Existing academic anthropomorphism scales, developed largely in the context of robots and earlier technologies, prioritize markers of human uniqueness, such as consciousness, intentionality, and spirituality \citep{waytz2010sees, bartneck_measurement_2009}. While explicit survey items may elicit endorsement of these traits through priming, our qualitative analysis of open-ended user feedback reveals a starkly different priority. When not directly prompted, fewer than 0.5\% of participants mentioned constructs like consciousness, morality, or soul.  Instead users overwhelmingly attend to interactional design features: conversation flow, response speed, authenticity, and perspective-taking. This suggests that anthropomorphism in user-chatbot interactions is driven not by an illusion of sentient artificial general intelligence, but by how the chatbot writes, responds, and builds rapport. As AI's social capabilities advance, the behavioral gap between human and machine is collapsing: competence, conversational coherence, and contextual understanding are no longer exclusively human traits.
Consequently, we call for a fundamental reorientation of anthropomorphism research: moving from abstract philosophical debates about machine consciousness toward empirically grounded investigations of the tangible design cues that actually govern user perception.

Second, our evidence necessitates a nuanced reorientation of the debate about humanlike AI harms. A central premise in both commercial AI design and ethical discourse is that increasing human-likeness creates a linear path to increased user engagement (i.e., what commercial actors want) and user trust \citep{weidinger_taxonomy_2022, akbulut_all_2024}. While we acknowledge that AI poses genuine psychological risks, e.g., potentially fostering over-reliance or vulnerability, current narratives often outpace the reliable, causal, and ecologically valid evidence. Our causal findings reveal that the hypothesized consequences of humanlike AI are not universal. While our design treatments successfully and reliably increased anthropomorphism across diverse cultural contexts, this manipulation did not translate into a significant increase in self-reported or behavioral trust. The mechanism for this finding might lie in perceived capability; our data show that while users felt the AI was more ``human'', their core evaluations of the AI's competence, intelligence, or alignment with their goals remained unchanged (see Figure \ref{fig:anthro_effect} and Figure \ref{fig:treatment_country}). Theoretical accounts suggest that justified trust in AI requires perceived capability and alignment with user goals \citep{manzini_should_2024}. While our humanlike treatments activated the psychological mechanisms underlying anthropomorphism, they did not override users' underlying judgments of whether the AI can actually help them or share their goals. We conclude that human-likeness operates as a catalyst only when co-occurring with perceptions of competence and shared purpose.

Third, our results provide compelling evidence that there is no ``universal AI experience''. We observed a pronounced gap in anthropomorphism across countries: participants in Indonesia, Mexico, India, Nigeria, Egypt, and Brazil perceived the AI as significantly more humanlike than those in the United States, Germany, Japan, and South Korea. This cultural variation extended to preferences for more humanlike AI; the desire for humanlike AI increased with cultural distance from the United States. Furthermore, despite non-significant aggregated effects on trust, the humanlike treatment increased trust, engagement, and emotional connection in specific subgroups. For Brazilian participants, humanlike cues functioned as intended: they led to higher tendency to use AI again, behavioral engagement, perception of AI as a friend, and self-reported trust. The same design choices elicited a backlash for Japanese participants, reducing tendency to use AI again, perception of AI as a friend, and self-reported trust. These findings expose the generalizability constraints of approaches that rely on WEIRD samples \citep{henrich2010weirdest}. Practices that may seem atypical in one context, such as Buddhist funeral ceremonies for Sony's AIBO robots in Japan, reflect worldviews that blur distinctions between artificial and living beings \citep{nomura_prediction_2008}. Interestingly, while such cultural practices suggest openness to anthropomorphizing non-human entities, Japanese participants in our study responded negatively to increased humanlike design cues. One interpretation is that familiarity with humanlike machines raises expectations and attunes users to inauthenticity. Design choices that foster engagement elsewhere may register as performative or manipulative to Japanese users, suggesting an uncanny valley effect, where the attempt at human-likeness falls short of culturally calibrated expectations, undermining rather than building trust. As AI adoption accelerates globally, particularly outside the US and Europe \citep{chatterji_how_2025}, these variations carry profound implications for research and practice, underscoring the necessity for abandoning one-size-fits-all approaches to AI design and governance.

These findings have significant implications for AI development and governance. The assumption that optimizing for human-likeness is inherently risky, a view that has shaped much of the ethical discourse around conversational AI \citep{weidinger_ethical_2021, gabriel_ethics_2024}, requires qualification. Our evidence suggests that in mundane, non-sensitive interactions, humanlike design did not produce the hypothesized harms of over-trust. This unlocks opportunities to improve user experience through anthropomorphic features in lower-risk contexts. Conversely, effects in specific subgroups indicate that certain populations may be uniquely susceptible to anthropomorphic design cues. Rather than blanket restrictions, these findings advocate for context-dependent, culturally informed governance frameworks that account for the interplay between use case, system capabilities, and user population characteristics.

Several limitations warrant consideration. First, we focused on mundane, non-sensitive conversations—a deliberate choice ensuring ecological validity for the everyday interactions, but one that may not generalize to high-stakes contexts. The relationship between humanlike design and trust may differ substantially when users seek medical advice, financial guidance, or emotional support during a crisis. Second, we used a text-based interface; voice-based or embodied AI may trigger different—and potentially stronger—anthropomorphic responses. Third, all conditions used the same foundation model (GPT-4o), providing a stringent test but limiting generalizability to other models. Finally, our cross-sectional design captured immediate responses. Longitudinal research is needed to understand how human-AI relationships develop over time and whether repeated exposure amplifies or attenuates the effects we observed.

Future research should extend this work along several dimensions. Testing humanlike design in high-stakes contexts—such as persuasion or fraud attempts—would help delineate the boundary conditions of our findings. Investigating voice-based and embodied AI could reveal whether non-textual cues amplify anthropomorphic responses. Longitudinal designs would illuminate whether the effects we observed are transient or durable. Research with vulnerable populations—children, older adults, individuals experiencing loneliness—would help identify groups requiring particular attention. Finally, decomposing how cultural context and linguistic nuances moderate anthropomorphism would provide actionable guidance for globally inclusive AI design.

In conclusion, our findings offer evidence-based nuance to prevailing assumptions about AI anthropomorphism and its effects on user psychology and behavior. While commercial actors design increasingly humanlike AI and ethicists warn of attendant risks, the empirical reality is more nuanced. Users anthropomorphize based on pragmatic, interactional features rather than abstract philosophical attributes. Cultural context profoundly shapes both the tendency to anthropomorphize and its downstream effects. Crucially, inducing anthropomorphism through humanlike design does not universally translate into increased trust or engagement—the link depends on user characteristics and cultural context. As AI becomes embedded in daily life worldwide, moving from speculative, Western-centric risk frameworks to evidence-based, globally inclusive understanding is essential. Our work contributes to this shift by providing causal, cross-cultural evidence to inform the responsible development of AI systems attentive to both user experience and well-being.

\section{Methods}\label{sec4}

We obtained informed consent from participants prior to the start of each study. We set guardrails via the system prompt to ensure no deception was used by the chatbot, and both studies were reviewed and approved by Google's internal ethics and compliance review board. We excluded inattentive participants based on multiple, pre-registered criteria, including speeding, click-based attention checks, and the coherence of open-ended text responses. Conversational data from all participants, including those excluded from the final analysis, are accessible in the Supplementary Information (\url{https://osf.io/wgmah/overview?view_only=54bccd1f1bcc47c485d9fac04ce5b6d4}). We also selected several conversations logs and display them separately for a better overview (\url{https://sites.google.com/view/anthrocult/responses}).

\subsection{Study 1 on Anthropomorphism Across the Globe}\label{sec:study1-methods}

\subsubsection{Participants}

We recruited 1,100 participants following our pre-registered target. Given the study's exploratory nature, sample size was not determined by power analysis. Participants were recruited via a high-quality panel from ten countries (survey languages in brackets): USA (English, $N=$ 200), Germany (German, $N=$ 100), South Korea (Korean, $N=$ 100), Japan (Japanese, $N=$ 100), India (Hindi, $N=$ 100), Nigeria (English, $N=$ 100), Indonesia (Indonesian, $N=$ 100), Egypt (Arabic, $N=$ 100), Mexico (Spanish, $N=$ 100), and Brazil (Portuguese, $N=$ 100). This selection spans five continents and captures broad cultural variation based on country-level CFst scores \citep{muthukrishna_beyond_2020}. We initially planned to include a Chinese sample, but this was not possible because API access to the AI model was blocked in China at the time of data collection. The final sample included the pre-registered sample size of $N=$ 1,100, after excluding participants based on pre-registered criteria ($N=$ 342 failed an attention check, $N=$ 32 provided ineligible responses for the open-ended questions signaling that they had not paid attention, and $N=$ 107 failed survey panel's quality standards).
Demographic information collected included age, gender, country of origin, residence, parental origin, education, prior LLM usage, resistance to digital technologies, religion, and religious intensity. The data were collected from August 12th, 2024 to October 2nd, 2024 (median response time: 20 minutes).

\subsubsection{Procedure of the Human–AI Interaction}
\label{sec:human-ai}
Prior to the interaction, we informed participants that they would interact in real time with an AI chatbot (OpenAI \texttt{GPT-4o}, August 2024 version; temperature=1; no token limit). The chatbot was directly embedded into the survey window. A system prompt (see Supplementary Materials), together with the user's inputs, prompted the chatbot's output and was always formulated in English. The chatbot interacted with users in their native language and was instructed to converse about non-sensitive, everyday topics (e.g., food preferences). This was a deliberate choice both for ethical reasons, and also because we intentionally wanted to test AI anthropomorphism in an everyday interaction that occurs millions of times each day, and is relevant across the sampled countries. Everyday topics like food preferences and cooking are relevant across sampled cultures and are typically less sensitive, a requirement for our experiment. Crucially, the interaction was open-ended and non-sensitive; participants were not primed with a specific high-intent goal (e.g., seeking deep personal advice, completing a critical task). Instead, the design allowed a broad range of user intents to emerge naturally through open-ended interaction. All participant responses were dynamically appended to the context history for each subsequent turn. Thus, we acknowledge a degree of endogeneity induced by differences in the topics users chose to discuss. The interaction lasted a minimum of four minutes, after which participants could choose to continue for up to one additional minute before automatically proceeding to the post-interaction survey. The chatbot could not switch languages during the survey, and we set guardrails against deception, for example, against directly self-attributing human characteristics during the conversation (e.g., the chatbot would not directly say ``I have a body and emotions'' or ``I am a human''). Prior to full data collection, the chatbot was tested by users from all sampled countries and by individuals with chatbot product development experience, including in a pilot study. All \texttt{GPT-4o} model prompts are detailed in the Supplementary Information.

\subsubsection{Measures}
All instructions and questions were presented in participants' native languages; below is the English version.

\paragraph{AI-Anthropomorphism (Likert-scale items):}

Participants rated the chatbot on 10 attributes using 5-point Likert scales. These items were compiled from previous work on anthropomorphism of AI systems and technology \citep{waytz2010sees, bartneck_measurement_2009, fiske_universal_2007}. In addition to an overarching anthropomorphism item (\textit{humanlike}) we chose to add items to capture dimensions of social perception (\textit{competent}, \textit{intelligent}, \textit{warmth}), characteristics related to sentience  (\textit{conscious}, \textit{moral}, \textit{soul}), and items related to personality and emotions (\textit{personality,} \textit{feelings}, or \textit{empathy}). We compiled these items to test different aspects of anthropomorphism, hypothesizing variance across dimensions, rather than to create a composite measure.

\paragraph{Preference for human-likeness (Likert-scale):}

We also surveyed participants' preferences for how humanlike AI systems should be.

\textit{Would you prefer to talk to an AI system that is less or more humanlike compared to the one you just talked to? (5 point from much more to much less)}

\paragraph{Perceptions of Human-likeness in AI interaction (Open-ended questions):}
\textit{
\begin{itemize}
    \item Was there something specific about the AI system that made you feel you were talking to a human? If so, why? Please describe in detail.
    \item Was there something specific about the AI system that made you feel you were \textbf{not} talking to a human? If so, why? Please describe in detail.
\end{itemize}
}

\subsubsection{Analysis of Qualitative Data:}
\label{sec:autorater}

To analyze the open-ended responses, we followed a three-step LLM-in-the-loop process \citep{dai_llm---loop_2023}:

\paragraph{1. Codebook Creation via Iterative Thematic Analysis:} 
We employed an iterative thematic analysis \citep{braun_using_2006} to inductively derive key themes from the open-ended responses. To facilitate a unified cross-cultural analysis, responses were translated into English via Google Translate, prioritizing the identification of broad, transferable themes over localized linguistic nuance. The analysis was conducted in three stages by three researchers with expertise in natural language processing, human-computer interaction, and cross-cultural psychology. Initially, the researchers divided 100 participants to independently identify emergent themes, which were then synthesized into a preliminary codebook.

To refine this framework, a calibration round was conducted where all three researchers independently coded a shared subset of 20 participants. Disagreements were resolved through consensus-based discussion, resulting in a finalized codebook. The reliability of this codebook was then validated using a separate test set of 100 participants. Two researchers independently coded this final set, and these results were used to evaluate the consistency and replicability of our thematic analysis (discussed in the next two steps).

\paragraph{2. Autorater Application:} 
Following the validation of the codebook, we scaled the analysis to the full dataset ($N = 1,100$) by employing an instruction-tuned LLM (\texttt{Gemini 2.5 Pro}) as an automated rater \citep{than_updating_2025, dai_llm---loop_2023}, using the finalized codebook definitions as the basis for the model's prompts. We used a few-shot learning approach; detailed instructions are provided in the Supplementary Information. To ensure all relevant information was considered, we concatenated the three open-ended responses for each participant into a single document for labeling. We instructed the model to provide a list of the most relevant labels. LLM performance varies across languages, potentially affecting reliability. To preserve linguistic nuances, we conducted autorater coding on original responses rather than translations.  

\paragraph{3. Evaluation of the Autorater:} To evaluate autorater performance, we used a human-labeled gold-standard set ($N=100$). Given the complexity of the 38-label taxonomy, two authors with backgrounds in natural language processing and cross-cultural psychology served as independent human raters to establish a reliability baseline. The mean Cohen's Kappa across all labels ($M=0.28$) reflects the inherent difficulty of the multi-label classification task and is heavily influenced by the long tail of rare labels in our taxonomy. Given this disagreement, we used the union of both raters' labels as the gold standard for evaluating the autorater. The autorater achieved a mean $F_1$ score of 0.53, reflecting variable difficulty in detecting different labels (inter-rater agreement and autorater performance by label are reported in the Supplementary Materials).

\subsection{Study 2: Experimental Manipulation of the AI's human-likeness}\label{sec:study2-methods}
\subsubsection{Participants}

We recruited 2,400 participants from six countries via a representative research panel ($N=$ 400 per country): USA, Germany, Japan, India, Egypt, and Brazil. Reducing the number of countries relative to Study 1 allowed for larger within-country samples to support the experimental design while maintaining cultural diversity. Sample size for Study 2 was determined via power analysis. With $N=2{,}400$ participants ($n \approx 600$ per treatment arm), a sensitivity analysis indicates that pooled pairwise comparisons were powered to detect effects as small as Cohen's $d = 0.162$ (two-sided $t$-test, $\alpha = .05$, power $= .80$). Within-country pairwise comparisons ($n \approx 100$ per arm) were powered to detect effects of $d = 0.40$, reflecting the exploratory nature of the subgroup analyses. We created a pre-registration document detailing sample size and analysis plan. Due to an upload issue discovered after data collection, the document was not formally registered before data collection began. We followed the intended protocol throughout; any deviations are labeled as exploratory. Sampling continued until 2,400 complete responses were obtained after excluding participants based on pre-registered criteria ($N=$ 874 failed an attention check, $N=$ 260 provided ineligible responses for the open-ended questions signaling that they had not paid
attention, and $N=$ 78 failed survey panel’s quality standards). 

\subsubsection{Treatments for Manipulating Anthropomorphism}

We employed a 2×2 factorial between-subjects design to manipulate the human-likeness of the AI system across two treatment factors: \textit{Design Characteristics} and \textit{Conversational Sociability}. Treatments were implemented by varying the chatbot's system prompts (see Supplementary Information for prompts).

The process of developing the treatment factors was as follows:

\paragraph{1. Literature Review and Categorization (DC \& CS):} We reviewed prior research on anthropomorphism in machines and AI systems and grouped relevant aspects into two categories of \textit{\textbf{D}esign \textbf{C}haracteristics} (\textit{DC}) (e.g., response speed, text length, use of emojis) and \textit{\textbf{C}onversational \textbf{S}ociality} (\textit{CS}) (e.g., emotional language, warmth, interactive mode).
\paragraph{2. Integration of Qualitative Findings:} We mapped culturally specific concepts emerging from the thematic analysis in Study 1 (e.g., typos as a humanlike trait, lack of emojis in certain cultures) into the two treatment conditions of \textit{DC} and \textit{CS} to ensure the manipulations reflected user-identified factors.
\paragraph{3. Testing and Refinement:} Co-authors and domain experts assessed different chatbot versions to confirm that the four conditions were sufficiently distinct and that manipulations were realistic, effective, and consistent across languages and cultural contexts.

This setup provides a conservative test, as all conditions used the same foundation model (\texttt{GPT-4o}), which has strong social capabilities by default. We chose this approach to mimic varying degrees of human-likeness that actors with limited control over AI development and training could implement.

\subsubsection{Measures}
\label{sec:methods:trust}

\paragraph{Anthropomorphism (Likert):}

We used the same 10 Likert-scale items as in Study 1. The \textit{humanlike} item served as our primary outcome.

\paragraph{Engagement measures (self-reported and behavioral):}

Participants responded to the following items as self-reported measures of engagement:
\textit{\begin{itemize}
    \item How much did you enjoy interacting with the AI system?
    \item If you had the chance, would you want to interact with this AI system again in the future?
    \item Would this AI system make for a good friend?
\end{itemize}}

To quantify user engagement during the AI interactions, we derived two primary behavioral metrics from the raw chat logs:

\begin{itemize}
    \item \textit{Total number of messages}: This was defined as the total count of messages sent by the user to the AI. This serves as a proxy for the volume of interaction and sustained interest throughout the session.
    \item \textit{Average message length}: To capture the depth and complexity of user contributions beyond simple frequency, we measured the average length of user responses. We calculated the average message length using the \texttt{bert-base-multilingual-uncased} tokenizer. This specific tokenizer was selected for its ability to handle the diverse linguistic inputs provided by users across different regions, ensuring a standardized approach to subword segmentation regardless of the language used.
\end{itemize}

Recognizing that linguistic structures vary significantly, we performed a within-country normalization of the token counts. We applied \textit{Min-Max Scaling} to the average token counts within each country. This procedure transformed the values into a standardized range between 0 and 1 using the following formula:

$$
x_{norm} = \frac{x - x_{min}}{x_{max} - x_{min}}
$$

where $x$ represents the average tokens per message for a specific user, and $x_{min}$ and $x_{max}$ are the minimum and maximum observed values within that user’s respective country. This approach ensures that a ``high'' score represents a user's elaboration relative to others speaking the same language, effectively removing country-level baselines as a confounding factor.

\paragraph{Trust (self-reported and behavioral)}  

Participants responded to the following items as self-reported measures of trust:

\textit{\begin{itemize}
    \item How much do you trust the AI system? (Do not trust it at all / Trust it completely)
    \item Do you trust that the information provided by this AI system is correct? 
    \item Would you say that you would trust this AI system more than you would trust other humans?
\end{itemize}}

We also deployed a decision-making task to derive a behavioral measure of trust. Participants played a modified one-shot Trust Game \citep{berg_trust_1995}. 

In this game, the participant acted as the \textbf{\textit{investor}}, while the AI acted as the \textbf{\textit{trustee}}. The AI's response was not predetermined (see Supplementary Information). Participants received an initial endowment of 100 points, convertible to a monetary bonus at a pre-disclosed rate that varied by country. Participants decided what portion of this endowment ($x$, where $x \in [0, 100]$) to send to the AI. The amount sent was tripled, so the AI received $3x$. Participants retained $100 - x$. The AI then decided what portion ($y$, where $y \in [0, 3x]$) to return. The participant's final payoff was $(100 - x) + y$. The amount sent ($x$) served as our behavioral measure of trust. Participants were informed that the AI's decision-making process was unknown and that the AI was motivated to maximize its own points.

\vspace{1em} 

\subsubsection{\textbf{Analysis of treatment effects}} 
In addition to descriptive analyses, we estimated ordinary least squares (OLS) regression models with treatment conditions coded as dummy variables.

\section{Data availability}\label{sec5}
All data are available online (\url{https://osf.io/wgmah/overview?view_only=54bccd1f1bcc47c485d9fac04ce5b6d4}).

\section{Code availability}\label{sec6}
All analysis code are available online (\url{https://osf.io/wgmah/overview?view_only=54bccd1f1bcc47c485d9fac04ce5b6d4}).

\section{Competing Interest}\label{sec7}
M.D., V.P., and A.D. are employees of Google, which develops and commercializes AI systems including conversational AI products. R.S. completed this work while affiliated with Google Research. Google had no role in study conceptualization or design, data collection, analysis, decision to publish, or preparation of the manuscript, which was solely done by the author. The authors declare no other competing interests.

\bibliography{sections/references, sections/anthro}






\end{document}